\documentclass[sigconf]{acmart}
\AtBeginDocument{%
  \providecommand\BibTeX{{%
    \normalfont B\kern-0.5em{\scshape i\kern-0.25em b}\kern-0.8em\TeX}}}

\copyrightyear{2024}
\acmYear{2024}
\setcopyright{rightsretained}
\acmConference[ICMI '24]{INTERNATIONAL CONFERENCE ON MULTIMODAL INTERACTION}{November 4--8, 2024}{San Jose, Costa Rica}
\acmBooktitle{INTERNATIONAL CONFERENCE ON MULTIMODAL INTERACTION (ICMI '24), November 4--8, 2024, San Jose, Costa Rica}\acmDOI{10.1145/3678957.3685728}
\acmISBN{979-8-4007-0462-8/24/11}

\usepackage{pifont}
\usepackage{multirow}
\usepackage{textgreek}
\usepackage[commandnameprefix=always]{changes}

\usepackage{caption}
\usepackage{subcaption}
\usepackage{hyperref}

\begin{document}

\title{Distinguishing Target and Non-Target Fixations with EEG and Eye Tracking in Realistic Visual Scenes}

\author{Mansi Sharma}
\email{s8mamisr@stud.uni-saarland.de}
\affiliation{%
  \institution{Saarland University, DFKI}
  \city{Saarbrücken}
  \country{Germany}
}

\author{Camilo Andrés Martínez Martínez}
\email{camilo.martinez@dfki.de}
\affiliation{%
  \institution{DFKI, Saarland University}
  \city{Saarbrücken}
  \country{Germany}
}

\author{Benedikt Emanuel Wirth}
\email{benedikt_emanuel.wirth@dfki.de}
\affiliation{%
 \institution{DFKI, Saarland Informatics Campus}
  \city{Saarbrücken}
  \country{Germany}
}

\author{Antonio Kr\"uger}
\email{antonio.krueger@dfki.de}
\affiliation{%
  \institution{DFKI, Saarland Informatics Campus}
  \city{Saarbrücken}
  \country{Germany}
}

\author{Philipp M\"uller}
\email{philipp.mueller@dfki.de}
\affiliation{%
  \institution{DFKI, Saarland Informatics Campus}
  \city{Saarbrücken}
  \country{Germany}
}

\renewcommand{\shortauthors}{Mansi Sharma, et al.}

\begin{abstract}
Distinguishing target from non-target fixations during visual search is a fundamental building block to understand users' intended actions and to build effective assistance systems.
While prior research indicated the feasibility of classifying target vs. non-target fixations based on eye tracking and electroencephalography (EEG) data, these studies were conducted with explicitly instructed search trajectories, abstract visual stimuli, and disregarded any scene context.
This is in stark contrast with the fact that human visual search is largely driven by scene characteristics and raises questions regarding generalizability to more realistic scenarios.
To close this gap, we, for the first time, investigate the classification of target vs. non-target fixations during free visual search in realistic scenes.
In particular, we conducted a 36-participants user study using a large variety of 140 realistic visual search scenes
in two highly relevant application scenarios: searching for icons on desktop backgrounds and finding tools in a cluttered workshop.
Our approach based on gaze and EEG features 
outperforms the previous state-of-the-art approach based on a combination of fixation duration and saccade-related potentials.
We perform extensive evaluations to assess the generalizability of our approach across scene types. Our approach significantly advances the ability to distinguish between target and non-target fixations in realistic scenarios, achieving 83.6\% accuracy in cross-user evaluations. This substantially outperforms previous methods based on saccade-related potentials, which reached only 56.9\% accuracy.

\end{abstract}




\begin{teaserfigure}
  \includegraphics[trim={0cm 0cm 0cm 0cm},clip,width=\textwidth]
  {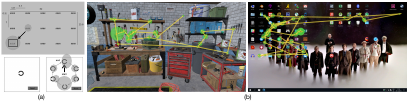} 
  \vspace{-0.7cm}
  \caption{(a) Previous work: Simplistic search displays with instructed search trajectory (b) Our work: Realistic search scenes with free search trajectory, search target: \texttt{Wrench, Outlook}} 
  \Description{(a) Previous work: Simplistic search displays (b) Our work: Realistic search scenes.}
  \label{fig:teaser}
\end{teaserfigure}


\maketitle

\section{Introduction}

Visual search is ubiquitous in daily life.
It occurs both when we search for items such as tools in a cluttered physical environment, but also in human-computer interaction when e.g. searching for a specific application icon.
Due to its ubiquity, assistance during visual search can potentially have a large impact on user satisfaction and task completion times~\cite{boot2009stable, guo2022effects}. 
For instance, an assistive system that knows that the user is currently fixating on the icon of an application they want to open can automatically launch this application for the user, overcoming the Midas touch problem in gaze-based interaction~\cite{elmadjian2021gazebar, mohan2018dualgaze}. 
On the other hand, a system that is supposed to help the user in locating a tool in a cluttered environment only needs to be activated when the user has not yet found the desired object.
The foundation for such novel assistive systems is the ability to distinguish fixations on the search target (\textit{target fixations}) from fixations that are not on the search target (\textit{non-target fixations}).

Previous work documented the feasibility to classify target from non-target fixations based on electroencephalography (EEG)~\cite{brouwer2013distinguishing}, as well as a combination of EEG- and eye tracking features \cite{brouwer2017eeg}, with a combination of saccade-related EEG potentials and a fixation duration feature being most promising.
However, these previous studies are limited in two key aspects:
First, they utilized simplistic abstract stimuli such as Landolt Cs and strings of symbols hidden by ('\#') 
on uniform backgrounds, disregarding any scene context.
Second, they did not let users search freely for the target, but explicitly instructed search trajectories.
This is in stark contrast to application scenarios where user search freely in a complex, structured scene such as in a workshop, or on a computer desktop.
Indeed, research in psychology indicates that scene context has an important influence on visual search~\cite{RN987, RN988,10.1145/985692.985780,RN992}.
It therefore remains an open question to what extent the results achieved with simple stimuli on uniform backgrounds can be transferred to more realistic stimuli placed in complex, cluttered scenes.

In our work we for the first time study the classification of target vs. non-target fixations from EEG and eye tracking recordings during free visual search in complex scenes.
In particular, we study two highly application-relevant scenarios: searching for a tool in a cluttered workshop, and searching for an icon on a desktop screen.
In a 36-participant user study we recorded gaze behavior and EEG measurements of people performing a visual search task in 140 different, diverse scenes from these two scenarios.
We present an automatic approach to distinguish target from non-target fixations based on state-of-the-art Common Spatial Pattern (CSP) EEG features
and a fixation duration feature. 
Our approach reaches 83.6\% accuracy in cross-user evaluation and clearly outperforms the approach based on saccade-related potentials used in previous work for target vs. non-target fixation classification (56.9\% accuracy)\cite{brouwer2017eeg}.
Furthermore, we perform extensive evaluations on the impact of scene type on classification performance.
Our results indicate that to reach maximum performance it is crucial to train on data from the same scene domain, and adding data of additional scenes during training did not lead to improvements.


\section{RELATED WORK}

The present work is related to studies investigating scene influences on visual search as well as to previous approaches to classifying target vs. non-target fixations.

\subsection{Visual Search and Scene Context}
Traditionally, cognitive research on visual search focused on highly-controlled, but artificial target and non-target stimuli that were presented in specific spatial configurations on blank backgrounds \cite{RN203, RN377, RN381}. However, starting in the late 2000s, an increasing number of studies investigated search for real objects in naturalistic scenes \cite{RN987, RN988, RN989}. This research argues that when viewing a scene, observers can quickly derive the "gist" of that scene via a non-selective pathway, that is, a pathway that does not rely on identification of individual objects \cite{RN575}. The information extracted from this scene gist, in turn, guides attention based on syntactic and semantic principles \cite{RN872, RN986}. Syntactic guidance refers to physical constraints that describe structurally plausible locations for specific objects. For example, if we assume that observers are searching for a hammer in the left scene depicted in Figure~\ref{fig:teaser}(b), the observers will likely not search for the hammer in front of the gray brick wall or in front of the red cupboard because the hammer would fall to the ground due to gravity. Instead, observers would more likely search on horizontal surfaces or on vertical surfaces where tools can be attached (like a pegboard). Semantic guidance refers to the meaningfulness of an object in a specific location. For example, observers would be more likely to search for the hammer on the pegboard or the workbench than on the floor. While it is physically possible for the hammer to lie on the floor, it is more likely that the hammer is stored on the pegboard or the workbench. In addition to syntactic and semantic guidance, contextual cueing enhances visual search by leveraging learned associations between targets and their environments \cite{pollmann2020preserved, geringswald2012simulated, geringswald2013contextual}. When observers repeatedly encounter similar contexts, they form implicit memories of these spatial layouts, allowing them to locate targets more quickly in future searches. For instance, repeatedly finding a hammer on a pegboard will cue the observer to search there more efficiently, demonstrating how learned context can optimize search performance.

Search for desktop icons can be considered an intermediate scenario that shows some characteristics of traditional visual search paradigms but other characteristics of visual search in naturalistic scenes: On the one hand, desktop icons are aligned according to an imaginary grid and the icons are artificial stimuli with only symbolic meaning. On the other hand, the icons are usually not presented on a blank background but on a scenic background as depicted in the right panel of Figure~\ref{fig:teaser}(b). Moreover, icons can also be grouped together by semantic principles (for example the icons of different web browsers or of different video games might be grouped together). Consequently, eye-tracking research has shown that both factors characterizing traditional visual search paradigms such as icon distinctiveness/complexity \cite{RN991} or inter-icon distance \cite{10.1145/985692.985780} as well as factors characterizing search within naturalistic scenes such as semantics \cite{RN992} play a role in visual search for desktop items.

\subsection{Target vs. Non-Target Fixation Classification}
While inferring the identity, category, or appearance of visual search targets has received some attention in research \cite{sattar2015prediction,kaunitz2014looking, hancock2013improving, barz2020visual,strohm2021neural}, few works addressed the task of whether a specific user fixation falls on a target or non-target \cite{brouwer2017eeg, brouwer2013distinguishing}. 
\citet{kamienkowski2012fixation} investigated electrophysiological responses to targets and non-targets in a visual search paradigm by using EEG epochs locked to fixation onset -- so-called fixation-related potentials (FRPs). More specifically, they compared FRPs in a visual search task where participants were allowed to freely scan the search display with event-related potentials (ERPs) observed in typical experimental designs where participants are asked to keep their gaze fixed (and targets and non-targets are presented in a temporal sequence). The authors showed that a typical difference between target and distractor-elicited ERPs also occurs for target and distractor-elicited FRPs (more specifically, a P300-like late effect at parietal electrodes). Based on this finding by Kamienkowski et al., \citet{brouwer2013distinguishing} proposed a method to classify target vs. non-target fixations by using FRPs. In their study, participants were presented with search displays containing six Landolt Cs arranged in a circular pattern, each with one of four possible orientations: a gap at the top, bottom, left, or right (see bottom Figure~\ref{fig:teaser}(a)). Importantly, participants were instructed to scan this search display in a particular, pre-defined order.
%
The findings revealed a mean accuracy of 62\%, as determined by Cohen's kappa, in classifying individual FRPs as either target or non-target among 11 participants.

In a follow-up work, \citet{brouwer2017eeg} employed a monitoring task where participants identified the location of the target in different cognitive load conditions. 
Their study setup presented a matrix (see top Figure~\ref{fig:teaser}(a)) to monitor 15 systems represented by a string of hashes (`\#\#\#\#').
Targets would randomly be activated, changing their state from `\#\#\#\#' to either `\#OK\#' or `\#FA\#'.
Participants had to indicate which systems failed (i.e. `\#FA\#') by pressing a button. 
The authors collected data from 21 users and created a balanced dataset with target- and non-target fixations.
Based on fixation duration and event-related potentials (ERPs) locked to the onset of a saccade (saccade-related potentials, SRPs) and an SVM classifier they achieved an average accuracy of 65\% in within-user evaluations.
While these previous works illustrate the potential of using EEG- and eye-tracking to classify target from non-target fixations, they are limited in two crucial aspects when considering realistic application scenarios.
First, the stimuli utilized in previous works were simple geometrical shapes or characters presented on a blank scene.
In contrast, we designed a large variety of realistic visual search scenes in two highly relevant application
scenarios: searching for icons on desktop backgrounds and finding tools in a cluttered workshop.
Second, participants did not search freely for the target, i.e. \citet{brouwer2013distinguishing} used a pre-defined search trajectory, and in \citet{brouwer2017eeg} participants were asked to always immediately fixate on the highlighted item.
In contrast, our study used a search paradigm where participants were allowed to freely make saccades and fixations across the whole image in order to find the target (as typical for visual search in everyday-life scenarios).


\section{Data Recording}
\paragraph{Participants} We recruited $36$ volunteers ($17$ female and $19$ male) aged between $19$ and $32$ years old (µ = $23.47$ , ~σ~=~$2.69$~). All participants reported normal or corrected to normal vision, and none had prior exposure to the study design. The study was approved by our institution's Ethics and Hygiene Board.
\paragraph{Hardware Setup} 
To display visual stimuli, we used a monitor with a resolution of $1920$ x $1080$ and screen brightness of $300$ cd/m2. 
We used the Enobio $20$ channel system\footnote{https://www.neuroelectrics.com/solutions/enobio/20} by Nonelectrics to record EEG signals with a sampling frequency of $500$~Hz. Electrodes were placed according to the $10$-$20$ international electrode placement system. 
To record eye-tracking data, we used wireless Tobii pro Nano\footnote{https://www.srlabs.it/en/scientific-research/hardware-products/tobii-pro-nano/} attached to the monitor's lower bezel with a sampling frequency of $60$ Hz. The distance from the screen was approximately 27.54$^{\circ}$ of visual angle.
The device was calibrated using 5-point calibration at the start of the experiment for each participant, using two coordinate systems. One is a $2$D system that spans the monitor with $(0,0)$ in the top right corner of the experiment setup monitor screen and $(1,1)$ in the bottom left. The second is a $3$D coordinate system for the experiment room, which measures the distance from the eye to the eye-tracker. We
synchronized eye-tracking, EEG and events of the
study procedure using labstreaminglayer \cite{LSL}.


\begin{figure}
  \includegraphics[trim={0cm 0cm 0cm 0cm},clip,width=\linewidth]{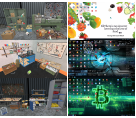}
  \vspace{-0.7cm}
  \caption{Examples of the Workshop and Desktop visual scenes we created for our study. In total, we created 140 unique scenes.}
  \label{scenes}
\end{figure}

\begin{figure*} 
  \includegraphics[trim={0cm 0cm 0cm 0cm},clip,width=\textwidth]{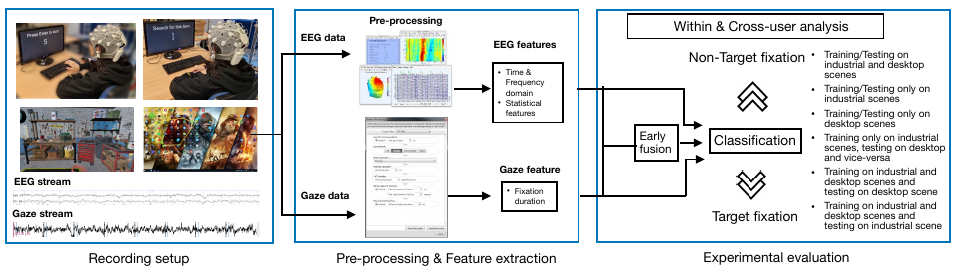}
  \caption{Outline of our approach to classify target from non-target fixations. We evaluate on seven experimental conditions for within and cross-user using EEG, Gaze, and Early fusion.}
  \label{Overview}
  \Description[Outline]{Outline of our approach to classify target from non-target fixations. We evaluate on seven experimental conditions for within and cross-user using EEG, Gaze, and Early fusion.}
\end{figure*}

\paragraph{Stimuli} 
We designed $70$ realistic scenes in Unity~\cite{unity} version \textit{2020.3.41f1} to simulate workshop environments, such as assembly units, manufacturing and production facilities, industrial labs, garage and repair workshop inspired by \cite{sharma2023implicit}.
In addition we synthesized $70$ desktop backgrounds with a wide range of differences in visual features like including texts/abstract patterns/human figures with diverse designs, colors, and motifs (\autoref{scenes} for examples) based on publicly available code by \cite{muller2022designing}. 
When constructing the workshop scenes, we depicted different levels of clutter by arranging machinery components, tools, workbenches, and other elements to resemble a typical workplace environment. 
To introduce different levels of complexity, we placed tools in various locations (inside cupboards, on the floor, and in unexpected areas) and orientations within the scenes. 
Our target stimulus could be one of five tools: Hammer, Pliers, Saw, Screwdriver, and Wrench. On average, each workshop scene comprised $250$ objects, with the minimum and maximum numbers of objects being $64$ and $455$, respectively. These numbers do not include the structural objects like walls, doors, furniture etc.
When designing the desktop scenes, we crafted diverse icons scattered throughout the background. 
We implemented a random selection method for determining the number of icons present. Our target stimulus could be an icon such as Google Drive, Facebook, Dropbox, Minecraft, Zoom, and many more. 
We used a unique target icon for each of the $70$ scenes. The number of icons placed in each desktop scene varied between a minimum of $50$ and a maximum of $97$, with an average of $73$ icons.
Each target was surrounded by an invisible bounding box which was later used to register participants mouse clicks on the target, as well as to define the ground truth of target- and non-target fixations.
This boundary provides a buffer of around 10 pixels, which helps to account for any slight inaccuracies in locating the target.

\paragraph{Procedure} 
Participants were presented with a general overview of the study. We showed them the setup and explained EEG and eye-tracking sensors to ensure they were comfortable tracking their physiological data. 
Next, we enabled participants’ to give their informed consent
and asked them to complete a demographics questionnaire. Participants were seated in a comfortable chair, and the distance between the user and the screen was $60$~cm. Next, the experimenter fitted the EEG cap on the participants’ heads, filling the electrodes with gel. Overall, the preparation time was about $20$ minutes. After visually inspecting the signal quality, we proceeded with the data recording, ensuring that electrode impedances remained below $15$ k$\Omega$ throughout the experiment. 
Our study design followed prior research~\cite{brouwer2013distinguishing, brouwer2017eeg} and introduced the target stimulus before prompting the search task. 
A single trial of the experiment

\noindent
consisted of $3$ steps. In Step $1$, we displayed the instructions for starting the study: 'Press enter to start', followed by a $5$~s counter. In Step $2$, we showed 
the target tool/desktop icon for $5$~s. In the last Step $3$,
the participant searches for the displayed target tool/desktop icon in the scene until the tool/icon is found. Figure~\ref{fig:teaser}(b) showed the workshop and desktop search scan path. Participants were instructed to click on the target using a mouse only after locating it. This minimized distractions with the mouse pointer and any influence of the hand movement artifact on EEG during the search phase. Participants were allowed to skip the search by pressing the 'S' key if they could not find the target.


The $140$ unique scenes were divided into four sessions, with breaks in between sessions. The initial two sessions featured workshop scenes, while the remaining two consisted of desktop scenes. The sequence of scenes and the target objects within each session were randomized. On average, the experiment lasted $120$~minutes per participant.


\section{METHOD}
In this section, we outline the pre-processing and feature extraction procedures for both EEG and eye-tracking data. We describe our experimental evaluation and baseline for both within-user and cross-user analyses. Figure~\ref{Overview} provides an overview of the data recording procedure, signal processing, and experimental evaluation conditions.

\subsection{Preprocessing}
\paragraph{Eye data pre-processing} Visual search tasks often rely on the analysis of fixations, which are the periods during which humans extract visual information~\cite{sattar2015prediction, borji2015eyes, sharma2023implicit}. We implemented the I-VT filter-based velocity-threshold fixation detection approach~\cite{olsen2012tobii} and used the default values of the parameters described below. Overall, we adhere to the seven-step approach outlined by~\cite{trabulsi2021optimizing}:
\textit{Gap fill-in} helps replace missing samples caused by participants blinking, looking away, or any other unforeseen disturbances causing short gaps in the data. \textit{Eye selection} averages the position data from the left and the right eye. We applied \textit{Noise reduction}, low-pass filtering to smooth out the noise while preserving the features using the moving median approach. We then used the \textit{Velocity calculator} to associate each gaze sample with a velocity. To classify the sample as either a part of fixation or not, we used the \textit{I-VT classifier}. Later, we applied \textit{Merge adjacent fixations} to correct erroneously split fixations due to noise. Lastly, using \textit{Discard short fixations}, we discard fixations that are too short to be relevant in visual search with a threshold of 60msec. 

\begin{table*}
  \caption{List of extracted features with PyEEG and added statistical features}
  \vspace{-0.3cm}
  \label{features}
  \centering
  \setlength{\tabcolsep}{40pt} 
\renewcommand{\arraystretch}{1} 
  \scalebox{1}{
  \begin{tabular}{lll}
    \toprule
    \textbf{Feature name}     & \textbf{Description}     \\
    \midrule
    Power spectral intensity & distribution of signal power over frequency \vspace{-0.1cm} \\ &  bands: delta, theta, alpha, beta, and gamma     \\
    \hline
    Petrosian Fractal Dimension   & ratio of number of self-similar pieces versus 
    \vspace{-0.1cm}  \\ & magnification factor      \\ \hline
    Hjorth mobility and complexity     &  mobility represents
    the proportion of the 
    \vspace{-0.1cm}  \\ & standard deviation of the power spectrum
    \vspace{-0.1cm}  \\ &  Complexity  represents the change in frequency       \\ \hline
    Higuchi Fractal Dimension     & computes fractal dimension of a time \vspace{-0.1cm}  \\ & series directly in the time domain        \\
    \hline
    Detrended Fluctuation Analysis & designed to investigate the long-range 
    \vspace{-0.1cm} \\ & correlation in non-stationary series      \\ 
    \hline
    Skewness & measure of asymmetry of an EEG signal  \\
    \hline
    Kurtosis & used to determine if the EEG data has peaked 
    \vspace{-0.1cm}  \\ & or flat with respect to the normal distribution
    \\
    \hline
    Minimum, Maximum, and Standard deviation   &  measure of variability of an EEG signal  \\
    \bottomrule
  \end{tabular}
  }
\end{table*}

\paragraph{EEG data pre-processing}
We cleaned the EEG data by applying high-pass filters with a cutoff frequency of $1$-Hz, along with a notch filter between $48$~Hz and $52$~Hz, and a low-pass filter with a cutoff frequency of $40$~Hz~\cite{klug2021identifying}. Following this, we removed any bad channels and interpolated them using the spherical interpolation method. We set the threshold for channel correlation rejection to $0.8$~\cite{delorme2004eeglab, 10.1007/978-3-031-30111-7_51}. Then, we referenced all channels to a common average~\cite{ludwig2009using}. To minimize correlation between electrodes, we performed independent component analysis using the Second Order Blind Identification (SOBI) algorithm~\cite{delorme2004eeglab}, followed by automatic rejection of components labeled as muscle, heart, and eye artifacts, using a $95$\% threshold. We epoched the pre-processed EEG data from the beginning of the target fixation through its duration, and did the same for the non-target fixation.


\subsection{Feature Extraction}
We utilized EEG and eye-tracking pre-processed data to compute feature sets from the target and non-target fixations in each trial, allowing us to customize our approach to obtain better results than the baseline.
\paragraph{Eye-tracking features}
Similar to the previous work by \citet{brouwer2017eeg}, our approach utilizes fixation duration, defined as the time that eye fixation was on the stimulus. As pointed out in the previous work, fixation duration is the most crucial feature for detecting targets compared to non-target fixations. In preliminary experiments, we utilized larger feature sets based on \citet{sharma2023implicit}, but did not observe improvements over fixation duration alone.


\paragraph{EEG features}
For EEG-based features, we utilized two different feature extraction approaches. Firstly, we used PyEEG - an open-source Python module for EEG feature extraction~\cite{PyEEG} that operates in the frequency and time domains. Additionally, we incorporated statistical features, resulting in a total of $15$ features per channel -

\noindent
similar to \cite{sharma2023implicit}. The EEG-based features and their corresponding calculations are described in detail in Table~\ref{features}. Secondly, we used Common Spatial Pattern (CSP) to extract features from EEG data, which maximizes the discriminative capacity of the features~\cite{sharma2023implicit, DBLP:conf/premi/ChatterjeeBKTKJ13, 10.1007/978-3-031-30111-7_51}. The spatial filter used in CSP is computed by solving a generalized eigenvalue problem. The resulting spatial filters distinguish between $2$ classes of EEG data and extract spatial patterns of brain activity associated with specific tasks. We used $15$ CSP features in addition to the default parameters from the MNE toolbox~\cite{GramfortEtAl2013a}.

\paragraph{Saccade-related potentials}
To compare our work with \citet{brouwer2017eeg}, we also adapted their saccade-related potential (SRP) approach. 
The original approach extracts EEG data for 1s starting from the highest velocity point of the saccade leading to the current fixation. 
With our eye-tracker's sampling frequency of 60~Hz, it was challenging to accurately determine the peak saccade velocity. 
As a solution, we opted to select the temporal midpoint of the saccade as the event onset point.

\subsection{Classification} \label{exp.ev}
Similar to previous work in fixation detection, we used Support Vector Machines (SVM)~\cite{brouwer2017eeg}.
We treated kernel as a hyperparameter and chose the best option from linear, polynomial, and radial basis function (rbf) kernel based on the validation set performance, along with $C$ values of (0.1, 1, 10) and $\gamma$ values of (0.1, 1, scale, auto) in the case of the rbf kernel.
We first performed a min-max scaling on the input features to harmonize the scale of features within- and across modalities. 
To fuse feature sets (eye-tracking and EEG features) we used early fusion, i.e. simple concatenation of feature vectors.


\begin{figure*}
\centering
  \includegraphics[trim={0cm 0cm 1cm 0cm},clip,width=\textwidth]{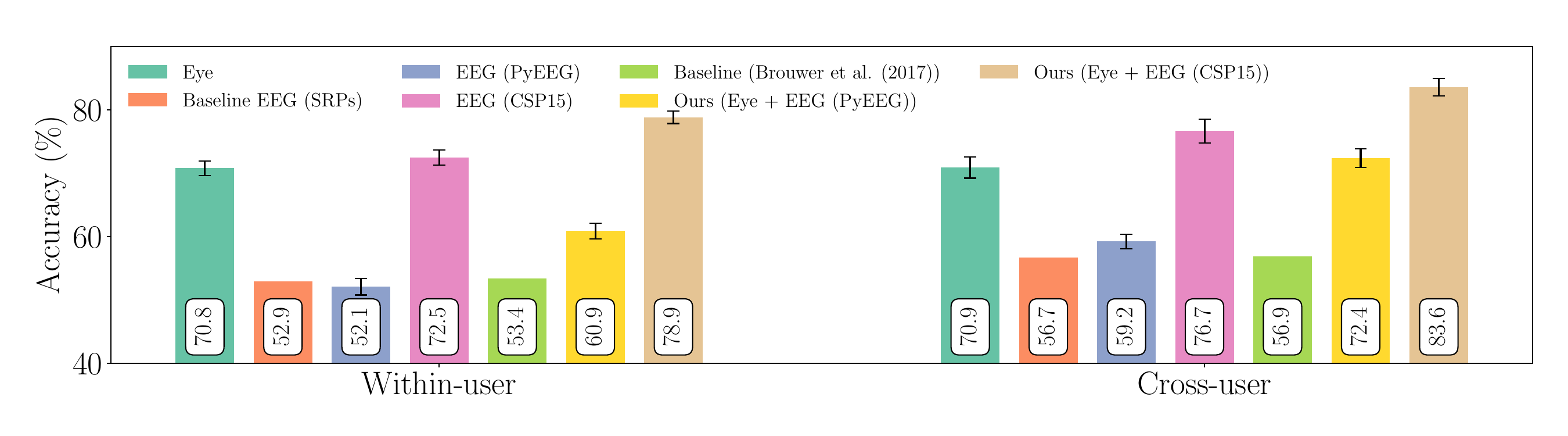}
  \caption{Comparison of within- and cross-user evaluations, training and testing on both workshop and desktop scenes. Error bars indicate $95\%$ confidence interval.}
  \label{res1}
  \Description[Outline]{res}
\end{figure*}   

\section{EVALUATION}
This section outlines the metric used to evaluate model performance and the experimental evaluation scenario. Additionally, we comprehensively describe our findings regarding specific scene domains and generalization across different scene domains.

\subsection{Ground Truth and Evaluation Metric}
We extract target fixations by collecting the first fixation inside the target bounding box for a given search trial.
All prior fixations in a given trial constitute the non-target fixations for this trial.
This procedure leads to a heavily unbalanced class distribution -- in line with previous work~\cite{brouwer2017eeg}, we randomly sub-sample the non-target fixation class to create a balanced class distribution on training and testing sets.
As a result, the random baseline is at 50\% accuracy in all evaluation scenarios. For each train-test split, we performed parameter tuning via grid search cross-validation only on the train data and used the test set only for prediction. 
In all evaluation scenarios, our evaluation metric is accuracy averaged over 10-fold cross-validation.

\subsection{Evaluation Scenarios}

We investigate a wide variety of evaluation scenarios to comprehensively test the generalization ability of our target vs. non-target fixation classification approach. 
While previous work only evaluated in within-user scenarios~\cite{brouwer2013distinguishing,brouwer2017eeg} (i.e. training and testing on the same user only), we also study cross-user prediction (i.e. splitting training and test sets by users). 
Furthermore, we study the generalization of our approach across scene domains.
For example, we evaluate to what extent a classifier trained on data from desktop images is able to classify target vs. non-target fixations on workshops, and vice versa.



\subsection{Overall Results}
In \autoref{res1} we present the result of our approach trained and tested on both scene domains (desktops and workshops).

Our approach based on Eye and EEG (CSP15) features achieved the best performance both in the within-user as well as the cross-user scenario.
In the within-user evaluation, our approach reaches the best accuracy of 78.9\%, clearly outperforming the method from \citet{brouwer2017eeg} (53.4\% accuracy). The fixation duration feature already reaches a performance of 70.8\% on its own, while the EEG (CSP15) features perform slightly better at 72.5\% accuracy. 
EEG (PyEEG) features perform far worse than CSP based features reaching 52.1\% accuracy. Interestingly, with fusion, results improve slightly leading to an accuracy of 60.9\%, but still far below the eye based fixation duration feature. Overall, CSP15 feature set seems to be very promising.

For cross-user evaluation, the pattern of results is very similar.
However, while the fixation duration feature reaches the same accuracy as in within-user prediction, the EEG-based features namely, CSP15 achieve higher accuracies. In line with the within-user evaluation, the PyEEG feature set performs worse, showing significant differences compared to our best-performing CSP515 feature set.
The best results are again obtained by our method with 83.6\% accuracy.

\begin{figure}
  \includegraphics[trim={1.5cm 0cm 3.5cm 2cm},clip,width=\linewidth]{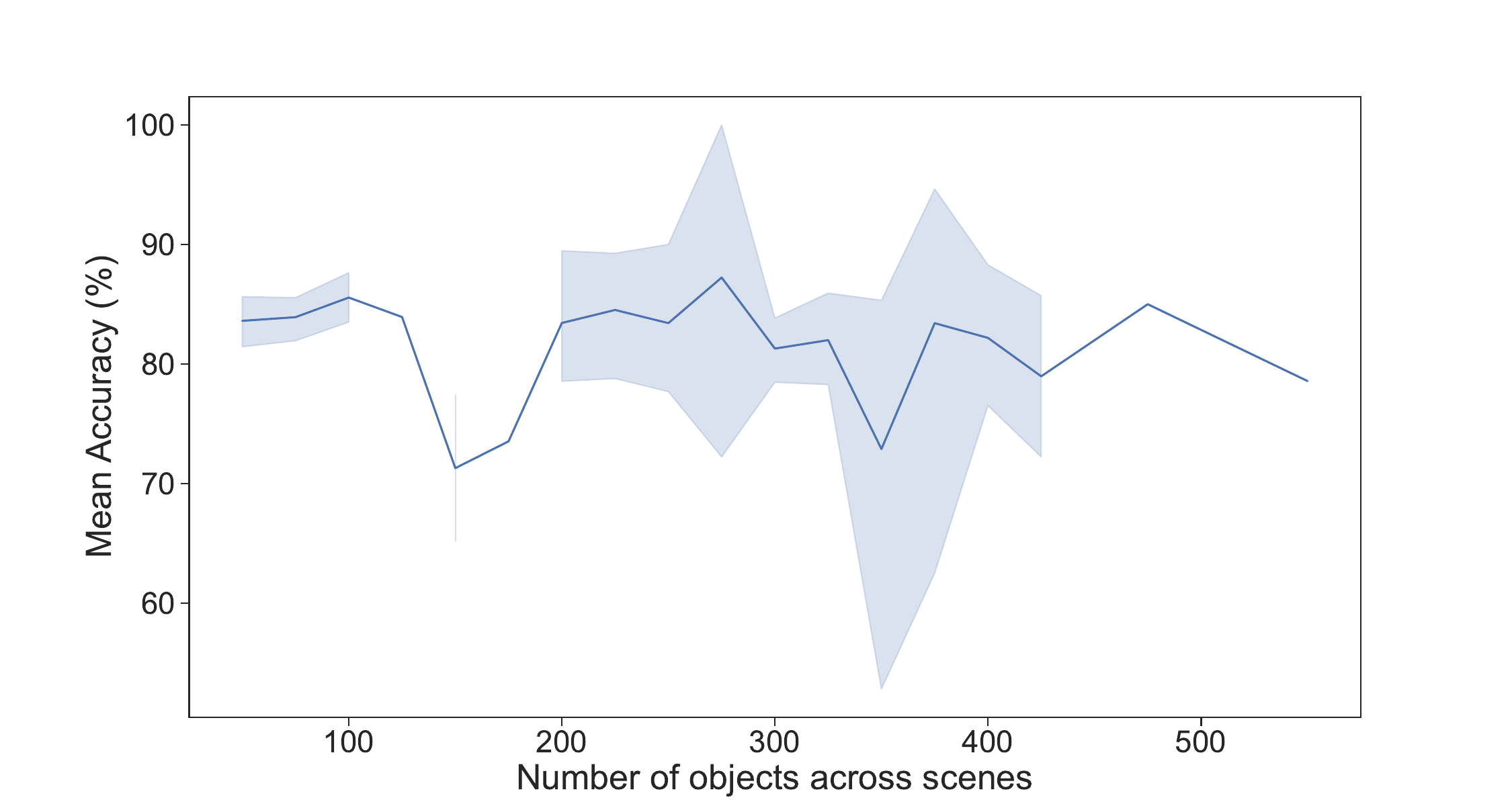}
  \vspace{-0.7cm}
  \caption{Effect of the number of objects on classification accuracy for the best performing case, cross-user with early fusion. The shaded region shows the confidence interval.}
  \label{scene_clutter}
  \Description{scene clutter}
\end{figure}

\autoref{scene_clutter} shows how the classification accuracy of a test set varies with the number of objects in a scene for the best performing case i.e. cross-user with early fusion. As the number of objects in the scene increases, the accuracy generally remains stable, highlighting the fact that there is no major impact of number of objects on the classification accuracy.

\begin{figure*}
\centering
  \includegraphics[trim={0cm 0cm 2.5cm 0cm},clip,width=\textwidth]{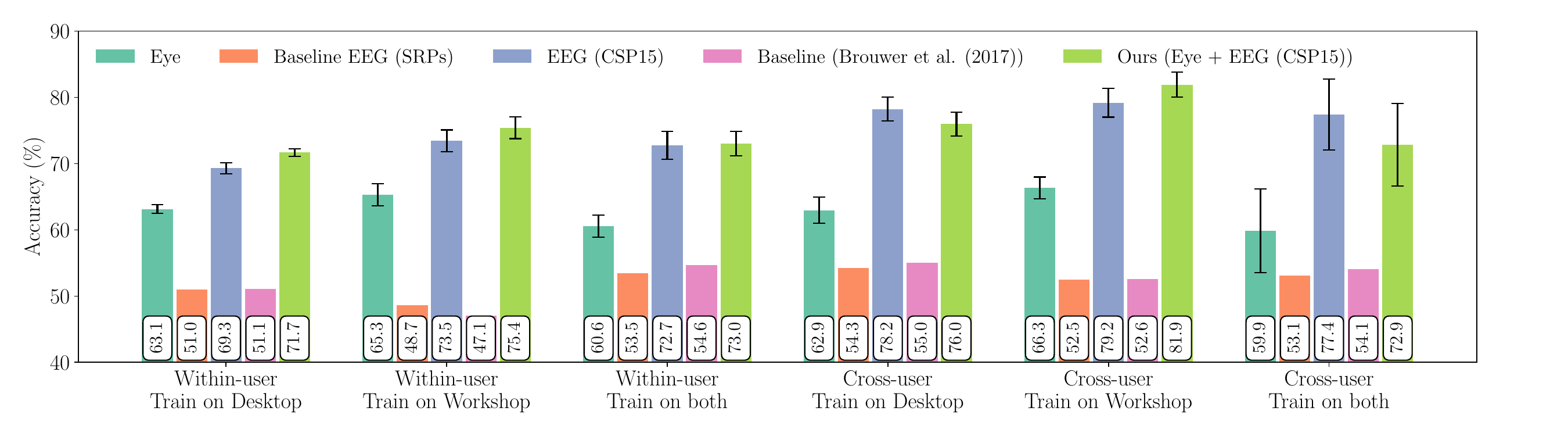}
  \caption{Comparison of within- and cross-user with best performing EEG features and testing only on workshop scenes using SVM. Error bars indicate $95\%$ confidence interval.}
  \label{res2}
  \Description[Outline]{res}
\end{figure*}   

\begin{figure*}
\centering
  \includegraphics[trim={0cm 0cm 2.5cm 0cm},clip,width=\textwidth]{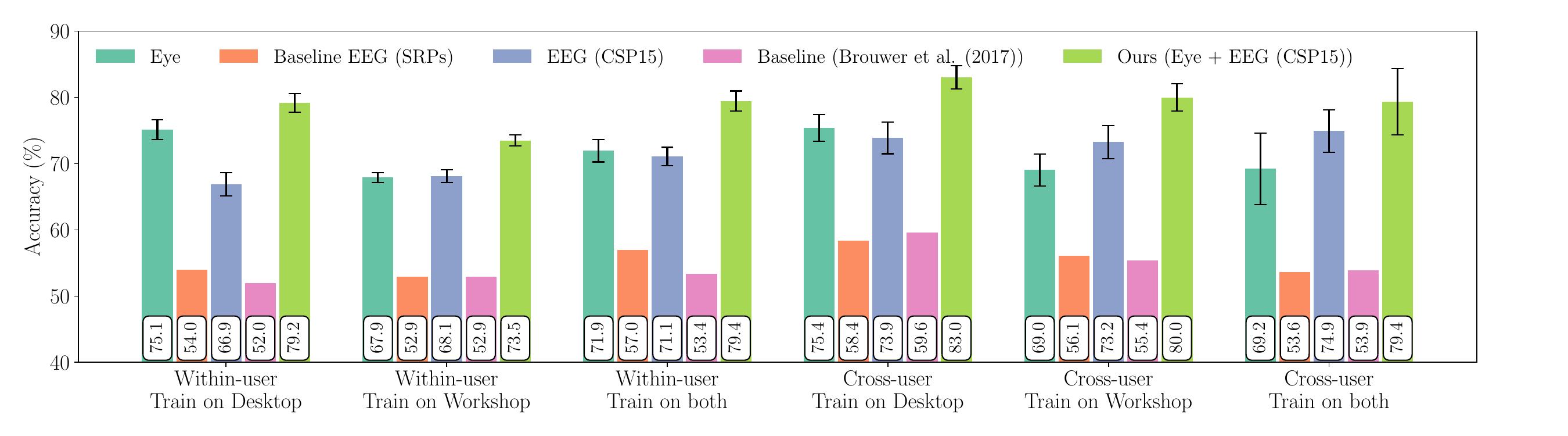}
  \caption{Comparison of within- and cross-user with best performing EEG features and testing only on desktop scenes. Error bars indicate $95\%$ confidence interval.}
  \label{res3}
  \Description[Outline]{res}
\end{figure*}

\subsection{Scene-Domain Generalization}

In the following, we present the results of our experiments across scene domains using the best performing EEG based CSP15 feature set. 

\paragraph{Testing on workshop scenes}
\autoref{res2} shows the results obtained with different training scenarios when testing on the workshop domain.
For both the within-user and the cross-user scenario, we investigate three different training setups: 
Training on workshop scenes (same domain), training on desktop scenes (domain transfer), and training on both industrial and desktop scenes. 
While the pattern of results is mostly similar to \autoref{res1}, we can make several interesting observations.
As expected, the same-domain evaluation scenario (training on workshops) achieves better results compared to the cross-domain scenario (training on desktops) in both within-user as well as cross-user evaluations.
However, also the cross-domain scenario still reaches accuracies far above the random baseline of 50\%.
The overall best accuracy in the within-user scenario is achieved by our approach trained on workshops with 75.4\% accuracy.
The best accuracy in the cross-user scenario is also achieved by our approach trained on workshops with 81.9\% accuracy.
Interestingly, training on both domains fails to clearly improve even over training on the desktop domain.

\paragraph{Testing on desktop scenes}
In \autoref{res3} we present the results when testing on desktop scenes.
Again, our method trained on the same domain achieved the best performance both in the within-user evaluation (79.2\% accuracy) as well as in the cross-user evaluation scenario (83.0\% accuracy).
In comparison to testing on workshop scenes, the eye feature achieves much higher performance, e.g. 75.4\% vs. 66.3\% accuracy for same-domain training in the cross-user scenario.
Furthermore, while EEG (CSP15) features consistently outperformed the eye feature when testing on workshop scenes, this is not always the case when testing on desktop scenes.
For example, in the same-domain training scenario (train on desktop) and within-user evaluation, EEG (CSP15) only reaches 66.9\% accuracy whereas the fixation duration features already achieves 75.1\% accuracy. 
Nevertheless, combining EEG (CSP15) features with the eye feature always results in clear performance improvements, highlighting the importance  of a multi-modal approach.




\section{Discussion}

In this section, we discuss the obtained results and focus
on our method’s performance and potential applications.

\subsection{On Performance}

Our approach to distinguish target- from non-target fixations based on EEG and eye-tracking achieved an impressive accuracy of more than 80\% in many evaluation scenarios, clearly outperforming the saccade-related potential approach from \cite{brouwer2017eeg}.
In line with previous work \cite{sharma2023implicit, blankertz2007optimizing, rekrut2021decoding} this underlines the effectiveness of the Common Spatial Pattern (CSP) features for practical EEG classification tasks.
Our results also indicate that the specificity of the training data in term of scene domain is crucial.
We consistently reached higher performance when training on the same scene domain compared to a different domain or even compared to extending the same-domain training set with an additional domain.
In contrast to many previous works on EEG signal classification~\cite{ma2022large, sharma2023implicit}, our EEG-based classifiers consistently reach higher performance in the cross-user than in the within-user evaluation scenarios.
One in important factor is of course training set size.
While training sets in the within-user scenario consist of $\approx$ 150-200 samples, in the cross-user scenario they are much larger with $\approx$ 6200-6400 samples.
An additional explanation for this result could also be that EEG signatures of target- versus non-target distinction are more person-independent compared to some other classification targets, such as e.g. emotions~\cite{zhao2021plug, sangineto2014we}.

\subsection{Applications}
The primary goal of our work is to contribute towards solving the Midas touch problem. Methods that can reliably distinguish between target and non-target fixations have numerous applications. For example, in the domain of human-computer interaction, these methods allow new ways for users to interact with their devices. Instead of using a mouse or touch pad in order to navigate through a file system, users could simply fixate on specific folders and files. Moreover, such methods can be used for efficient guidance systems: 
If the system is able to detect the target fixation of the user, it can understand the user's intentions and provide guidance for the next step. For instance, if the user is searching for a particular item on an e-commerce website, once the system detects that the user has found the desired item, it can suggest other related items or offer discounts on the purchase. This not only enhances the user's experience but also helps the business by increasing sales. 
Another potential application for such systems would be in the domain of forensic investigations. Cognitive research has invested much effort into the development of methods that allow to identify search targets that the user wants to conceal. For example, suspects that are presented with a line-up might search the line-up for their accomplices but keep it to themselves if they found any. Methods that allow another person to identify these concealed targets are subsumed under the term \textit{concealed information test} \cite{10.3389/fpsyg.2012.00342} and typical approaches are based on users' response times, eye movements or electrophysiological responses. Thus, a system that is able to integrate these measures to successfully detect even concealed search targets would substantially improve the reliability of the concealed-information test.

\subsection{Limitations and Future Work}

In our paper, we for the first time investigated the classification of target versus non-target fixations in free visual search in realistic scenes.
As it is the first study of its kind, it naturally has a number of limitations which need to be addressed in future work.
While we chose two highly application-relevant scene domains, namely desktop scenes and cluttered workshops, a larger variety of scenes will be needed in the future to more thoroughly understand scene influences and to build models that generalize better across scenes.


Another limitation concerns the practicability of our approach. Despite progress in the development of easy-to-use EEG systems (dry electrodes, mobile amplifiers, etc.), these systems still lack the signal-to-noise ratio that stationary systems with wet electrodes provide. However, preparing EEG recordings with the latter set-ups can be time-consuming and tedious, thus reducing user acceptance and adoption. 

While our approach based on the analysis of data from a single fixation already reached a convincing performance, we suspect that results can be improved by modelling the temporal aspects of search behaviour adequately.
While we did investigate a LSTM-based approach receiving input from several previous fixations we were not yet able to outperform the single-fixation approach presented in this paper.
As a result, we leave temporal modelling of the fixation sequence for target- vs. non-target classification to future work.
Furthermore, exploring and comparing different fusion methods like late and hybrid fusion and regression based analysis could provide new possibility to enhance performance \cite{vortmann2022multimodal, dimigen2021regression}. By experimenting with various strategies to combine data from multiple sources or features, we may discover more effective ways to classify fixations. Future research should focus on how these methods impact performance and consider combining them with temporal modeling to achieve better results.

Finally, our user study was screen-based.
While this is perfectly adequate for the desktop scene domain, the workshop scenes are an approximation of a real-world situation.
We took great care to construct realistically looking environments to elicit natural search behavior, but a domain gap to the real world will always be present when using the computer screen as a display device.
Future work should investigate target- vs. non-target classification in closer-to-real-life scenarios.
A first step will be virtual reality interactions, but ultimately a data recording in a physical environment will be needed to evaluate the domain gap.


\section{CONCLUSION}
In this work, we presented the first study on the automatic classification of target- versus non-target fixations during free search behavior in realistic visual scenes.
We conducted a 36-participant user study in which participants searched for target objects within scenes from two application-relevant domains: cluttered workshops and computer desktops.
We presented a classification method built on eye tracking and EEG features and showed clear performance improvements over previous approaches that were developed in artificial and highly-controlled search scenarios.
Our results highlight the importance of a match between scene domains at training and test time.
This underlines the importance of taking scene into account when building systems to distinguish target- from non-target fixations.

\section*{Acknowledgments}
This work is funded by the German Ministry for Education and Research (BMBF) under grant number 01IS17043 for Software Campus 2.0 (DFKI) and partially by grant number 01IW20003. Philipp M{\"u}ller and Benedikt Emanuel Wirth are funded by the European Union Horizon Europe programme under grant number 101078950.


\newpage
\bibliographystyle{ACM-Reference-Format}
\bibliography{sample-sigconf}

\appendix

\end{document}